\documentclass[letterpaper, 10 pt, conference]{ieeetran} 

\IEEEoverridecommandlockouts

\usepackage{epsfig} % for postscript graphics files
\usepackage{mathptmx} % assumes new font selection scheme installed
\usepackage{times} % assumes new font selection scheme installed
\usepackage{amsmath} % assumes amsmath package installed
\usepackage{amssymb}  % assumes amsmath package installed
\usepackage{graphicx}
\usepackage{textcomp}
\usepackage{xcolor}
\usepackage{caption}
\usepackage{subcaption}
\usepackage{multirow}
\usepackage{booktabs}
\usepackage{siunitx}
\title{\LARGE \bf
Haptic Communication in Human-Human and Human-Robot Co-Manipulation}

\author{Katherine H. Allen$^{1}$, Chris Rogers,  Elaine S. Short
\thanks{*This work was supported by the National Science Foundation }% <-this % stops a space
\thanks{$^{1}$All authors are with Tufts University, Medford, MA, USA
        {\tt\small kat.allen@tufts.edu}}%
}

\newcommand{\BibTeX}{\rm B\kern-.05em{\sc i\kern-.025em b}\kern-.08em\TeX}

\begin{document}

\maketitle
\thispagestyle{empty}
\pagestyle{empty}

%%%%%%%%%%%%%%%%%%%%%%%%%%%%%%%%%%%%%%%%%%%%%%%%%%%%%%%%%%%%%%%%%%%%%%%%%%%%%%%%
\begin{abstract}

When a human dyad jointly manipulates an object, they must communicate about their intended motion plans. Some of that collaboration is achieved through the motion of the manipulated object itself, which we call ``haptic communication.''  In this work, we captured the motion of human-human dyads moving an object together with one participant leading a motion plan about which the follower is uninformed. We then captured the same human participants manipulating the same object with a robot collaborator. By tracking the motion of the shared object using a low-cost IMU, we can directly compare human-human shared manipulation to the motion of those same participants interacting with the robot. Intra-study and post-study questionnaires provided participant feedback on the collaborations, indicating that the human-human collaborations are significantly more fluent, and analysis of the IMU data indicates that it captures objective differences in the motion profiles of the conditions.  The differences in objective and subjective measures of accuracy and fluency between the human-human and human-robot trials motivate future research into improving robot assistants for physical tasks by enabling them to send and receive anthropomorphic haptic signals.
\end{abstract}

%%%%%%%%%%%%%%%%%%%%%%%%%%%%%%%%%%%%%%%%%%%%%%%%%%%%%%%%%%%%%%%%%%%%%%%%%%%%%%%%

\begin{figure}[h!t]
    \centering
    \begin{subfigure}[b]{.7\linewidth}
    \centering
\includegraphics[width=\textwidth]{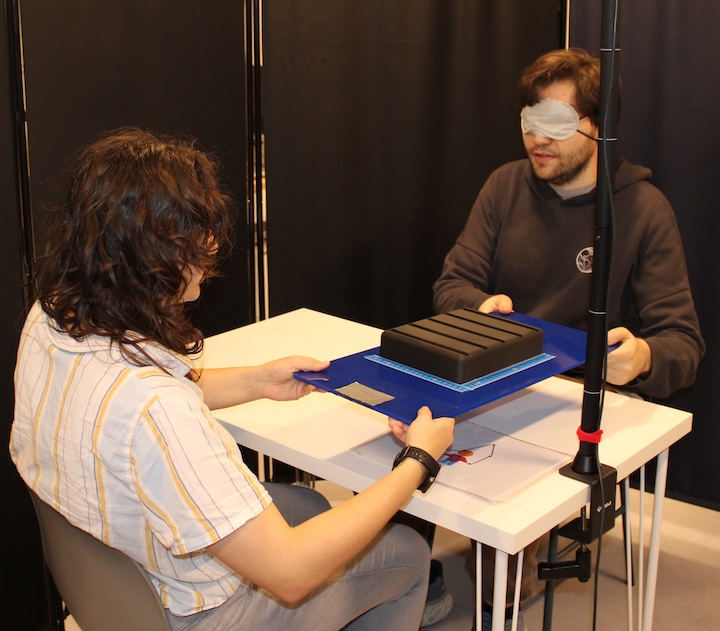}
    \caption{Human-Human study setup}% The Follower participant is blindfolded so that they can not see the motion plan cards.}
    \label{fig:participantsetup-hh}
    \end{subfigure}
    \hfill
\begin{subfigure}[b]{.7\linewidth}
    \centering
\includegraphics[width=\textwidth]{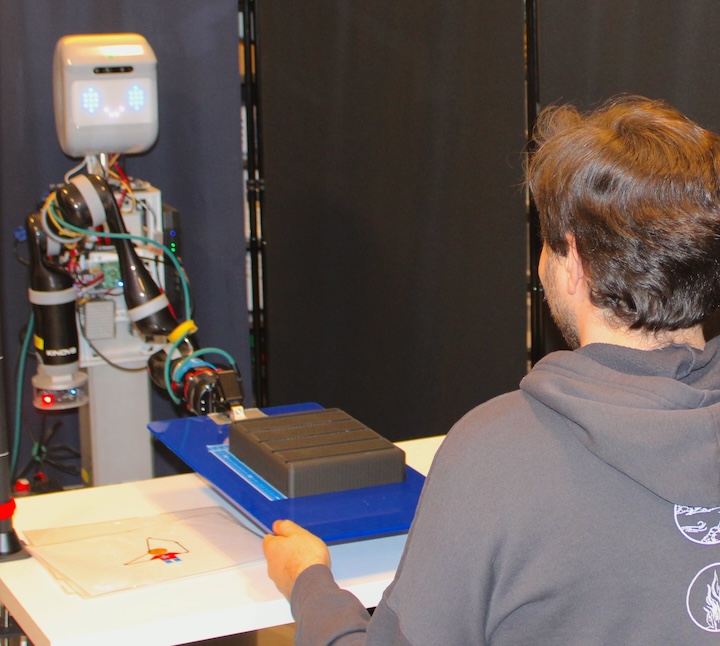}
    \caption{Study setup for the Robot Leading Human condition.}%  In the Human Leading Robot Condition, the robot would be blindfolded and the participant would not be blindfolded.}
    \label{fig:participantsetup-hr}
    \end{subfigure}
    \caption{Experiment Setups}%  In each part of the experiment, the Leader and Follower are across a small table from one another, and human participants are seated.}
    %\Description{Two images. In the left image, two people are seated across a small table, holding the plastic tray above the table.  One participant wears a grey blindfold.  In the right image, a participant is seen from the back seated at the same table, holding the same tray, but now with the mobile manipulator robot.}
\end{figure}

\section{Introduction}
In physical collaboration tasks like carrying a couch or moving a table, haptic signals are an important channel of communication between participants to coordinate the group action. In human-human interactions, the communication and interpretation of these signals is primarily subconscious, but prior research suggests that they may enable more efficient human-robot collaboration \cite{reed_physical_2008}. In order for robots to participate in this haptic conversation, we need to develop a more robust understanding of how haptic communication occurs in both human-human and human-robot interaction.  This knowledge can then be used to develop models for interpreting haptic intent, provide robots with comprehensible and predictable behavior, and avoid unwanted oscillations in collaborative manipulation.

In this paper, we present a study of haptic interaction, without use of visual or auditory signaling, during the collaborative manipulation of a shared object.  We compare human-human and human-robot dyads to test whether there are observable differences in the subjective fluency of human-human and human-robot dyads, and whether these correlate with changes in the character of acceleration profiles of the co-manipulated objects.  We additionally collect data on human perceptions of robot collaborators to identify potential co-variables in subjective fluency.

We conducted a user study with 34 participants.  In the study, two agents collaborated to move a shared object, with one participant designated as the motion leader and one as a follower. Each participant acted in one these roles, and interacted with both another human participant and a mobile manipulator robot.  We collected measures of subjective and objective measures of task fluency, as well as video and IMU recordings.  We found that the acceleration data from an IMU mounted on the shared object changes more smoothly in human-human dyads than in human-robot dyads, with more fluent collaborations having smaller accelerations overall and smaller changes in acceleration during the task.  We further find that common objective measures of collaboration fluency (e.g. task duration) do not correlate linearly with subjective fluency measures, and propose alternate measures based on our data. This work contributes to our understanding of the differences and similarities in current human-human and human-robot haptic communication during collaborative manipulation,  and provides insights that can inform future methods for autonomous haptic signaling by robots.

\section{Background}
Prior work in human-robot and human-human co-manipulation suggests the existence of some kind of difference between na{\"i}ve motion planning strategies and those actually used by human collaborators.  Reed et al \cite{reed_physical_2008} showed that a human-human dyad was able to complete a physical collaboration task more quickly than any of: the human alone, the human with a robot partner, or the human with a robot partner whom they believed to be a human partner. However, the human who mistakenly believed that they had a human partner completed the task more quickly than in the case where they knew their partner was a robot.   This suggests that there is a \emph{conversational} aspect to human-human haptic communication, beyond moving the object and waiting for the partner to catch up. ``Hesitation'' in the human response to a robot's ``least-jerk'' model lifting of a shared table-like object \cite{parker_design_2012} also suggests that a follower is waiting for some sort of indication from the leader of their plan.  Outside of haptic interaction, research has shown that motion patterns that are ``legible'', i.e.,  that allow the collaborator to infer an unknown goal based on the observed motion, reduced coordination and task time and improved subjective robot acceptance measures \cite{dragan_effects_2015}.   More recent work has also explored physical human-human interaction to classify collaboration styles from haptic features \cite{al-saadi_novel_2021,rysbek_recognizing_2023}, and identifying implicit communication in collaborative transportation scenarios by ``via subtle signals encoded in velocities transmitted to the transported object'' 
 \cite{yang_implicit_2025}, using them as a method for probabilistically identifying the user's goals based on the joint action.  Our work measures the acceleration of the co-manipulated object to identify whether these legible, conversational interactions are observable with a low-cost IMU, which could be used to train future models for robot co-manipulation control.

Research in developmental biology provides potential insight into what bio-memetic haptic communication might involve.  For example, pointing and joint attention gestures develop out of reaching behaviors\cite{carpendale_development_2010}, and objects are grasped differently according to their planned use \cite{rosenbaum_cognition_2012}. Trajectory modification is also sometimes used in joint manipulation grasping tasks to increase the probability of comprehension \cite{pezzulo_human_2013}.  In our study, we explore the motion of co-manipulated objects by human dyads to identify factors relevant to these signals---if robotic agents can identify human co-manipulation signaling, they can predict the planned motion and react accordingly. 

%A variety of classification algorithms have been explored to characterize haptic \emph{social} gestures (e.g. tap, rub, squeeze) using pressure sensors \cite{jung_automatic_2017}, or the built-in force-torque sensor of a standard robotic arm \cite{bianchini_towards_2021}, and substantial research has been performed into characterizing and classifying \emph{objects} via their haptic properties \cite{tatiya_deep_2019,sinapov_learning_nodate,chu_robotic_2015}.   In the control domain, simulated variable-impedence controllers in one DoF provide a controllable system  where the intention of the human participant is measured using the force applied to the object at the human's control point \cite{aydin_new_2014}.  Human study experiments have verified the need for wrist proprioception in human manipulation tasks \cite{parker_design_2012, kosuge_mobile_2000, ikeura_optimal_2002} suggesting that robot wrist-joint force detection may provide sufficient feedback for a controllable system in vertical lifting (1 DoF) scenarios.  Robotic systems with more sensors, such as arm force/torque data and friction-sensing gripper pads, have been used for rotational manipulation of a shared object\cite{gienger_human-robot_2018} with required contact/grip changes.  \fixme{does it?} Our work shows that a 6-DOF force-torque sensor is sufficient to detect the differences between human and robot motion leading behavior, and can characterize haptic communication via the force data transmitted to the robot arm.

Prior research has shown that timing of inputs is critical to human comprehension and to a stable human-in-the-loop control system.   In work with human-robot handoffs, a delay introduced during a handover improved compliance\cite{admoni_deliberate_2014}. Other work identifies an ideal 500ms period for object handoff \cite{chan_grip_2012}, and that control inputs for collaborative action occur at no more than 0.5 Hz\cite{parker_experimental_2011}.  In addition to meeting the human input criteria, the human-robot interaction must not induce instability into the underlying control system\cite{aydin_computational_2020}. Our research shows that, in addition to the timing of inputs, the character of the inputs (especially their magnitude and rate of change) is important to the perceived fluency of the interaction.
\section{Methodology}
We identify three hypotheses concerning the fluency and characterization of co-manipulation tasks:

\begin{figure}
    \centering
    \includegraphics[width=3in]{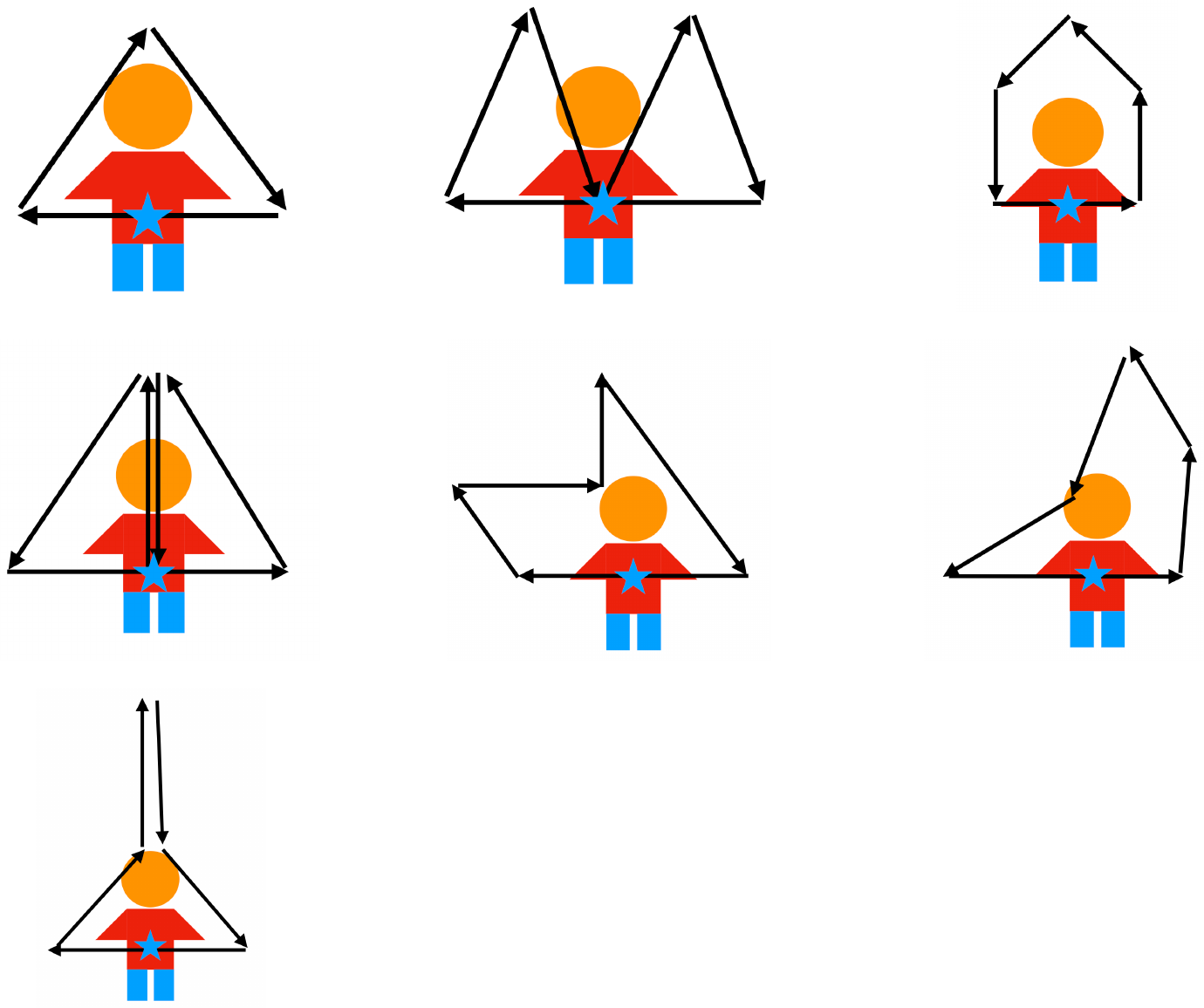}
    \caption{Motion Plan Cards for all experiment stages.  The triangle, 3-part motion plan card (top-left) is used for participants to practice the procedure before data is taken. The other 6 cards, each of which has a 6-stage motion plan,  are used for data collection, with three cards in each condition (human collaborator or robot collaborator.}
    \label{fig:motionplancards}
  %  \Description{Seven small images. Each has a geometric figure suggestive of a person, with an orange circle head, red rectangle with two attached triangles as a torso, and two blue rectangles as legs.  In front of each of the figures are a series of arrows creating a closed shape.  One shape is an equilateral triangle with the point up and a blue star bisecting the bottom arrow. The other six each have six segments, creating either a clockwise or counterclockwise circuit.}
\end{figure}

\begin{itemize}
    \item \textbf{H1:} Human-robot co-manipulation is less fluent than human-human co-manipulation. Subjective collaboration fluency will be higher for human-human interactions than for human-robot interactions    
    \item \textbf{H2}: There are ``haptic communication previews'' in recorded IMU acceleration data---changes in the acceleration of the co-manipulated object that occur before changes in the direction. 
    \item \textbf{H3}: These occur more frequently in collaborations that are subjectively described as fluent. 
  %  \item \textbf{H4:} In a human-robot co-manipulation, human leaders try to treat the robot like another person.  Human participants who rate the robot more positively in terms of social intelligence will have more fluent collaborations with the robot.  
\end{itemize}

In addition, we conducted an exploratory analysis of potential objective metrics for fluency by correlating them with the subjective fluency measurements from participant questionnaires:
\begin{itemize}
    \item{Task Completion Time}
    \item Average acceleration magnitude and change in acceleration experienced by the shared object/IMU (in all trials)
\end{itemize}

To investigate these hypotheses, we conducted a two by two between by within study, with a between-subjects condition of role (leader vs follower) and within subjects condition of partner type (robot vs human).  Each human participant was randomly assigned as either a leader or a follower, and collaborated with both a human and a robot partner.  Participants were additionally randomly assigned to experience either the robot partner condition first or the human partner condition first.  Table \ref{tab:experimentsequence} summarizes the conditions of the study and order randomization, with participants acting as either \textbf{L}eader or \textbf{F}ollower, and with a \textbf{H}uman partner or a \textbf{R}obot Partner first.  

We measured the fluency of collaboration using survey questions after each collaboration trial, to capture the participant's immediate reactions to the experience.  We collected an additional subjective measurement of fluency after all three collaborations in the condition were complete.

\begin{table}[h]
\caption{The four possible sequences for a human participant as either \textbf{L}eader or \textbf{F}ollower, and with a \textbf{H}uman partner or a \textbf{R}obot Partner, and the number of participants who experienced that sequence.}
    \centering
    \begin{tabular}{c|c|c}
& Leader & Follower \\
\midrule
Human First & 7 & 10 \\
Robot First & 10 & 7 \\
    \end{tabular}
    \label{tab:experimentsequence}
\end{table}

\subsection{Experimental Apparatus}
 At the start of the session, the Leader was given 3 motion plan cards selected from a set of six similar motion plans (Figure \ref{fig:motionplancards}).   The Leader additionally was given a card with a simplified motion plan card (a triangle) for practicing the protocol.   The cards used were pre-randomized as a balanced set: three were used in each of the two conditions (human partner/robot partner) such that all participants saw all the cards from the set during the experiment, in a random order.

In order to isolate the haptic communication channel from the effect of visual and aural communication, all participants were asked to remain silent during the experiment except to coordinate with the researcher.  In addition, participants in the Follower role were blindfolded during the manipulation of the object
to avoid accidental visual exposure to the motion plans.   Participants taking the Leader role were not blindfolded in order to enable them to use visual feedback for their own motion to replicate the trajectories on the cards.

\begin{figure}
    \centering
\includegraphics[width=0.45\textwidth]{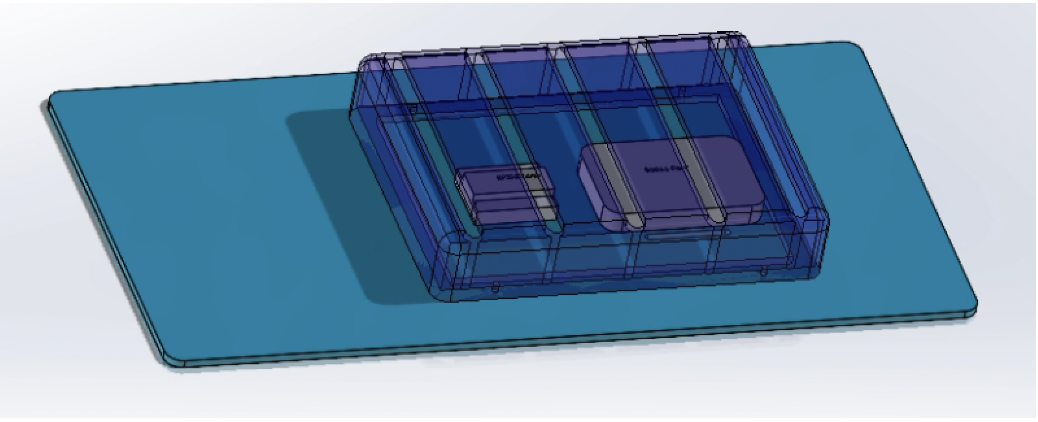}
    \caption{A transparent CAD view of the shared object, showing the acrylic board, 3D printed electronics cage, RP2040 Connect and LiPo battery }
    \label{fig:sharedobject}
    %\Description{A CAD drawing export of a blue board with rounded corners.  Filling the center half of the board, though slightly off-center, there is a transparent black box with stripes of vent holes across the top. Through the transparency, two smaller rectangles are visible with very small writing indicating that they are the battery and RP2040 board.}
\end{figure}

The shared object (Figure \ref{fig:sharedobject}) used in this study is a custom-made, low-cost, instrumented board approximately 30 cm wide, 60 cm long and .5 cm thick.  An Arduino RP2040 Connect running a custom web server controls the 2040's onboard Inertial Measurement Unit, giving 6 DoF data on the object's motion at its approximate center of mass. IMU data from the shared object and overhead video of the task execution provided ground truth data on the motion of the object as the participants manipulated it through the challenges.  

\begin{figure}[h]
\centering
\includegraphics[width=0.3\textwidth]{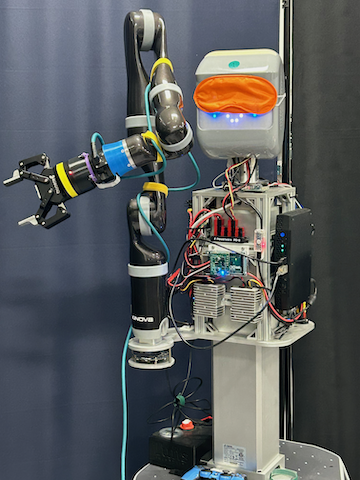}
\caption{The Social Mobile Manipulator Robot, blindfolded as in the Human Leading Robot condition. 
 %Participants were introduced to the robot by name, given a short explanation of what it means to be a mobile manipulator robot (e.g. that the robot can move around the floor and can manipulate things with their arm), and then the robot introduced themselves verbally and demonstrated the motion of their arm by moving into the ``start'' position. 
 The blindfold covers both the LED eyes and the robot's RGBD camera to strengthen the similarity in the mind of the participant between the robot and the participant's human collaborator in the other condition.}
 \label{fig:beepboop}
%\Description{An image of the mobile manipulator robot.  It has a low-resolution LED face on a grey plastic head, with two blue eyes and a pink mouth, mostly covered with an orange blindfold that also obscures the robot's RGBD camera. Below the head is a torso with many exposed wires. In front of the torso is a black Kinova arm, with several brightly-colored 3D printed pieces connecting a Bota SenseOne Force-Torque sensor and then a Robotiq gripper. A teal cable runs from the Bota SenseOne towards the body of the robot and is secured with bright yellow cable ties.}
\end{figure}

In the human-robot interactions, we added to the experimental apparatus a social mobile manipulator robot (Figure \ref{fig:beepboop}).  We used a social mobile manipulator robot rather than just a robot arm based on the  Reed's finding \cite{reed_physical_2008} that users appeared to use haptic signaling only when they believed that they had a human partner: by making the robot collaborator more anthropomorphic we increase the chances of haptic communication in the human-robot collaborations.  The robot took the place of one of the human participants in the collaboration, as either Leader or Follower.  

The social mobile manipulator robot used in this study is a 1.5m tall, custom-built humanoid robot, with a Fetch robotics base, central pillar and LED expressive face, a 7DoF Kinova Jaco 2 arm, a Bota SensOne force-torque sensor with IMU, and a Robotiq two-fingered 85mm gripper. It uses AWS Polly for speech synthesis of phrases in English to greet and say goodbye to participants using pre-generated phrases.
 
In the Follower role, the robot used a custom, hand-tuned proportional and derivative controller over the Cartesian forces detected at the force-torque sensor.  The sensor is located just above the gripper, to capture off-axis forces that are not reflected in joint torques due to system compliance.  The controller used the same proportional and derivative gains ($K_P = -0.4$, $K_D = -5.0$) applied in all three cardinal directions (x,y,z) and included a deadband of $\pm2$ Newtons for noise reduction.

In the Leader role, the robot used waypoints collected to match the trajectories from the motion planning cards and then processed into trajectories for robot arm movement using the OMPL planner for the Jaco2 arm. 

\subsection{Study Procedure}
We recruited 34 participants in 17 pairs from the university and the nearby community to collaborate in manipulating a shared object through a series of predetermined motion plans. Participants ranged in age from 18 to 54 years old, with 14 women, 18 men, and 2 non-binary people.   The protocol of this study was approved by the University's Institutional Review Board (IRB) as Study 00002967.

In each condition (human-human, robot leading human, human leading robot), the agent acting as the leader was given three of the motion plan cards, each with a plan for how to move the shared object, according to one of six trajectory-following motion plans (Figure \ref{fig:motionplancards}). To control against the follower pre-generating their own motion plans (which might agree or conflict with those of the leader in confounding ways), the follower in each condition did not have any prior knowledge of the motion plans and was blindfolded for the duration of the task.  
% Opportunity for future work: compare the difference between a follower having knowledge of the task and not having knowledge of the task, and whether their motion plan (if they had one) interfered with the leader's plan/communication.

We evaluated the collaborations both qualitatively and quantitatively, using annotated video recordings of participants and annotated motion data from the RP2040 Connect IMU on the shared object.  IMU data allows the capture of subtle motion characteristics which might be obscured in a purely visual observation.% and allows comparison between the motion previews and the full motion as executed. 

%At the end of all three tasks in the condition, participants were asked to rate overall collaboration fluency.

\subsubsection{Human-Human Collaborative Manipulation}
In the human-human collaborations, two humans participate in a physical collaboration task.  We record their motion via the acceleration data from the RP2040 Connect IMU.  Participants were asked to introduce themselves to their collaborator prior to the experiment, and to disclose their level of familiarity with their assigned collaborator, which is considered as a potential co-variable. %but not balanced within the experiment due to the complexity it would introduce in study recruitment.  
Each participant was assigned either the ``Leader'' role or the ''Follower'' role for all their interactions.

\subsubsection{Human-Robot Collaborative Manipulation}
In the human-robot collaborations, a human and an anthropomorphic robot participate in a physical collaboration task.   At the beginning of each participant's interaction with the social mobile manipulator robot of Figure \ref{fig:beepboop}, they were introduced to the robot and given a short description of its capabilities. %(e.g. that it is a mobile manipulator robot, which means that it can move around and can manipulate things with its arm), and the robot introduced itself verbally and demonstrated the motion of its arm by moving into the start position.  

Each of the 34 participants experienced the same condition (Leader or Follower) with the robot collaborator that they experienced with their human collaborator, but with a different subset of 3 cards to ensure motion plan novelty.  Half of the participants collaborated with the robot before collaborating with a human partner, and half experienced the collaboration with a human partner first, as per Table \ref{tab:experimentsequence}. 

Time sequences for the robot collaboration were captured both with the IMU data from the shared object and with force data and joint positions captured via ROSbag from the robot.

%At the end of both robot conditions, the robot thanked the human collaborator for participating in our study and said goodbye, and then moved into a resting position.  The participant then either completed the post-study questionnaires or moved on to the human-human portion of the study, depending on their assigned condition (Table \ref{tab:experimentsequence}).

\subsection{Measures}
The same subjective measures are used to evaluate the fluency of task collaboration for both the human-human and the human-robot interactions. %Additional subjective measures evaluate the robot's perceived social intelligence and the participants' attitudes about robots. Objective measures of fluency are tested against subjective measures to validate their effectiveness.

\subsubsection{Subjective measures of Task Fluency}
After each task, the participants were was asked to rate  %``How well do you feel your human collaborator followed your motion plans on this trial?'', and the motion follower was asked ``How well do you feel you understood and were able to follow your human collaborator’s motion plans on this trial?'', 
the fluency of the collaboration on a 1-5 scale.  These comprise the ``Task Rating'' of fluency.  In addition, after each set of three tasks with a single collaborator, each participant was asked to give an overall rating for fluency with this collaborator, the ``Overall Rating'' of fluency. 

% things I did not/can not go back and do: take notes or record  of participants' comments (no video sound)
% things I will probably not do: annotate fluency via the video data

\subsubsection{Task Duration Video Analysis}
Task duration is captured via video annotation and cross-checked with IMU and ROSBag datastamps.  In addition to task duration, the duration of each of the six legs of the motion plan (henceforth ``subtask durations'' ) were collected from visual observation of the video data. Video data was not captured for five interactions (four human-human and one robot-leader interaction), which were excluded from this annotation and subsequent analyses that rely on task or subtask durations. 

\subsubsection{Motion Characterization}
The IMU in the RP2040 captures acceleration in the X, Y, and Z directions and rotational acceleration around the X, Y, and Z axes, at a variable data rate between 2 and 5 Hz.   The IMU start time is written to the IMU data file at the beginning of data collection, and task start time is captured via video data.   In the version of the IMU data collection used in the study, the clock time at the end of the IMU recording was not recorded in the IMU data file, so the video-captured task end clock time is used as the end time for both.  

To account for the differences in data rate, task completion times, and wait time (the time between the start of IMU data recording and the video-annotated task start time), we scale the collected IMU data in two ways depending on the needs of our analysis:  

%\textbf{Scaling Method 1:  IMU data scaling 0-100 }
%The first method of scaling the IMU data simply scales the entire set of collected data from 0-100.  Breakpoints and wait time are also scaled such that they can be plotted for each dataset.

\textbf{Scaling Method 1: IMU data scaling to clock time }
The wait time is added to the task completion times (from video data), and then IMU datapoints are scaled to cover this clock time.   In this visualization, the variance in total task duration is visible between trials.

\textbf{Scaling Method 2:  IMU data scaling by task subsegment}
To more accurately compare each task subsegments and identify common features at the subsegment level, after the data was correlated to clock times, the subtask durations (identified by video annotation) were used to identify the breakpoints in each IMU file correlating to the point where the interaction leader changed their primary motion direction to follow the new leg of the motion plan.   

These were used to re-scale the data again, such that the duration of each subtask was equal to 1 unit.  This allows different participants' performances of the same motion plan card to be directly compared, even if they did not perform the plans at the same speed or at consistent speeds.   In this visualization, all breakpoints occur at the same time on the graph.

\textbf{Acceleration Data Calibration}
In addition to the following three scaling methods for the x-axis, we have normalized the y-axis values on each plot by taking the average of the first half of the wait time data for each and subtracting that from the full y-axis set, making the y-axis not acceleration but change from baseline acceleration.  This allows us to account for miscalibrations in the IMU that would make comparing trials more difficult, and re-zeros the z-axis acceleration to account for gravity.

\begin{figure}
    \centering
    \begin{subfigure}[t]{0.5\textwidth}
    \includegraphics[height=2.5in]{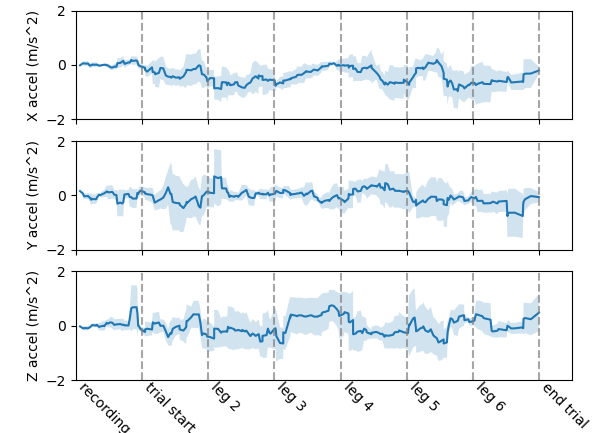}
    \caption{Human-human Trials of the DT card rated Fluency 5}
    \label{fig:humanhumanfluency5}
    \end{subfigure}
    \begin{subfigure}[t]{0.5\textwidth}
    \includegraphics[height=2.5in]{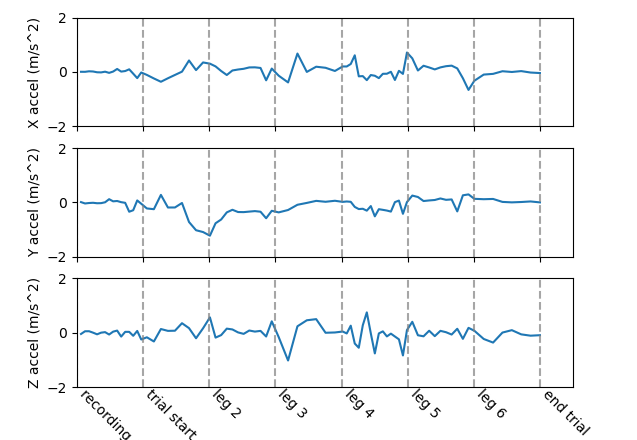}
   \caption{Robot Leading Human Trials of the DT card rated Fluency 5}
   \label{fig:robotleaderfluency5}
    \end{subfigure}
    
        \caption{Linear Acceleration Data in the x (left and right), y (toward or away from the participants), and z (up and down) directions, averaged over multiple trials}
        \label{fig:meanaccel}
\end{figure}

\textbf{Mean Time Series Acceleration by Card Type}
To identify changes in acceleration data that are most relevant to the motion plans associated with each card, we average the time series data by interaction type (human-human, robot-leading-human, or human-leading-robot).  For this analysis, an example of which is included in Figure \ref{fig:meanaccel}, we first scale by segment (Scaling Method 2), and then combine data into a single dataframe by timepoint.  For each time, we fill forward the acceleration data from the previous true measurement, which can then be used to average across the trials despite the data rate for each trial being inconsistent due to the time-scaling process.  

%\subsection{Participant Attitudes about and Perceptions of Robots}
%Prior to the experiment, we used the GAToRS \cite{koverola_general_2022} validated question set to evaluate participants' positive and negative personal and social attitudes about robots in general.   At the end of the experiment, a subset of the Perceived Social Intelligence scales \cite{barchard_measuring_2020} were used to evaluate participants' evaluations of the robot used in this study in terms of its recognition, adaptation, and prediction of human behaviors and cognition (Subscales RB, RC, AB, AC, PB, PC), as well as the robot's trustworthiness, rudeness, and helpfulness (Subscales TRU, RUD, and HLP, respectively).  These provide potential covariables for robot leader and follower behavior and evaluations of the robot's performance in the study tasks.

\section{Results}
\subsection{Fluency by Collaborator}
We measure subjective fluency with the overall rating given by each participant to their collaborator at the conclusion of the three trials, as well as with the rating given after each trial. 

\begin{figure}
% generated with python3 ~/Documents/GitHub/hapticcomm/processing/likertprocessing.py /Users/katallen/Dropbox/Research/HapticComm/Analysis/export-allcsv/fluencybycollaborator-overall.png 
\begin{subfigure}[t]{0.5\textwidth}
    \centering
    \includegraphics[height=2in]{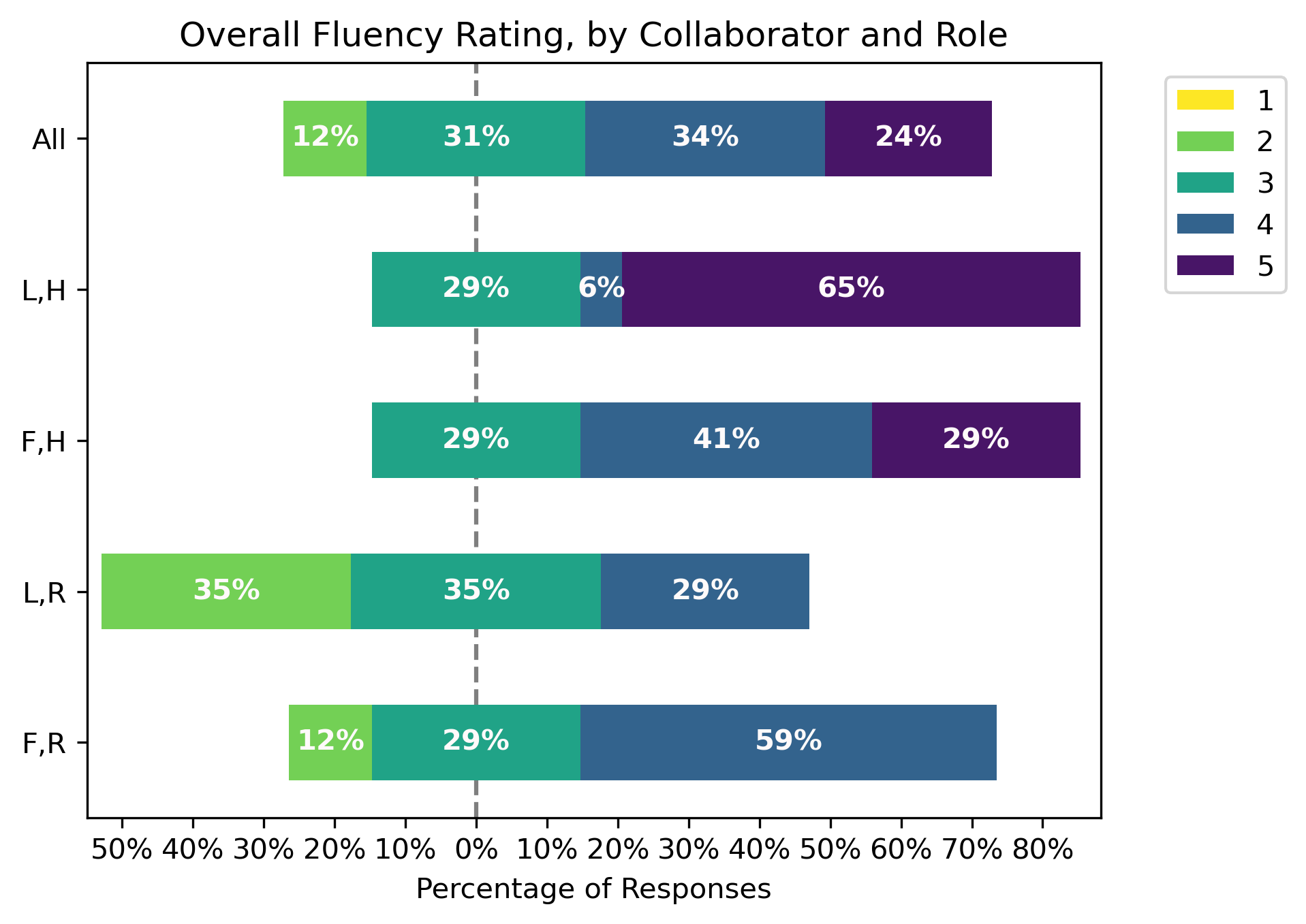}
    \caption{Overall Fluency (3 trials)}
    \label{fig:fluencybycollab-overall}    
\end{subfigure}
\begin{subfigure}[t]{0.5\textwidth}
    \centering
    \includegraphics[height=2in]{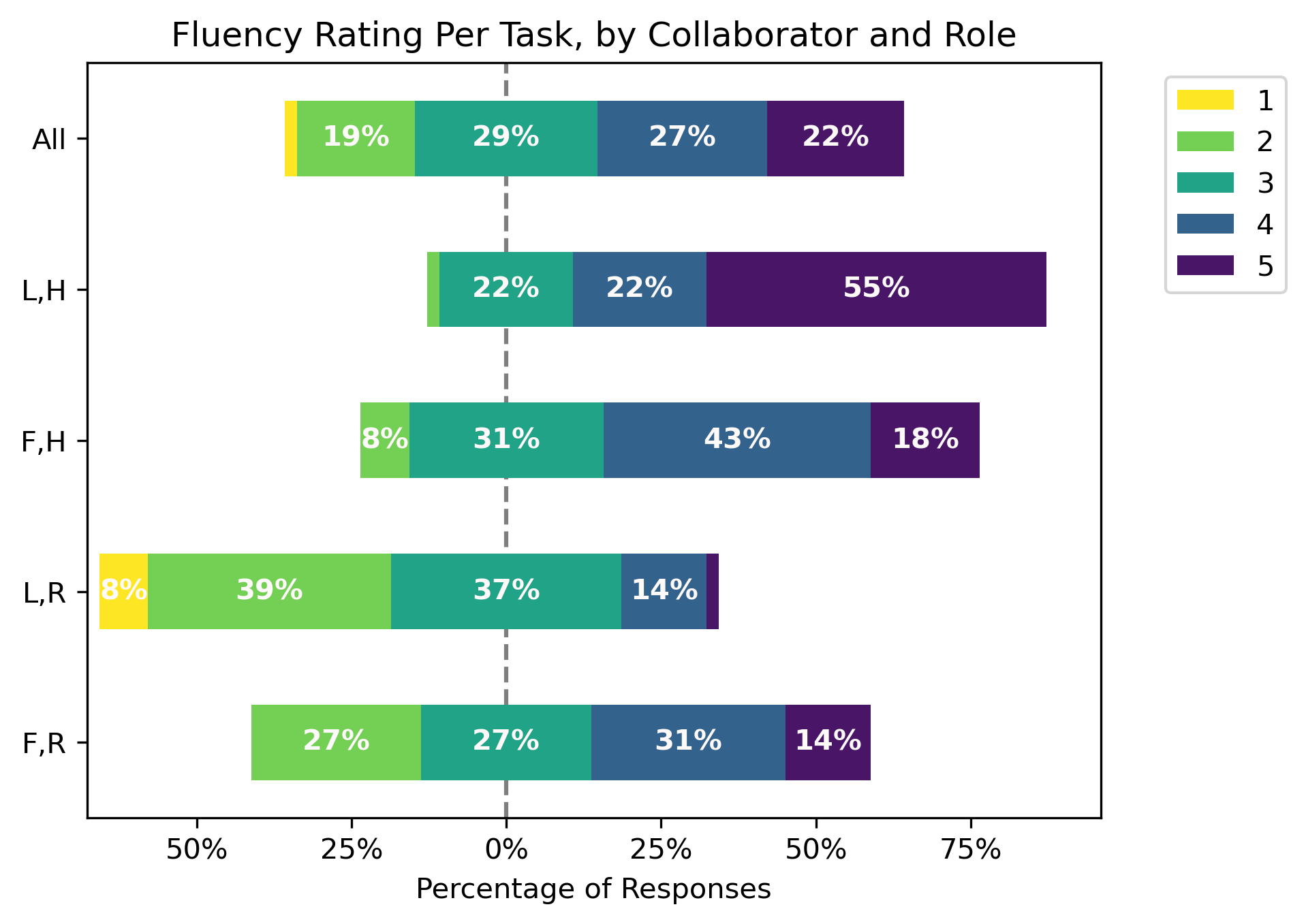}
    \caption{Task Fluency}
    \label{fig:fluencybycollab-trial}    
\end{subfigure}
    \caption{Fluency (1-5) by role (leader, L or follower, F) and collaborator (human, H or robot, R)}
        \label{fig:fluencybycollaborator}  
\end{figure}

Figure \ref{fig:fluencybycollaborator} and Table \ref{tab:fluencyratings} show the ratings given to human and robot collaborators.  Human-human interactions are rated as the most fluent overall, supporting hypothesis \textbf{H1}.  Leaders are more likely to rate the collaboration as fluent than followers. 

%Fluency ratings did differ between participants in the same human-human interaction, but are not straightforwardly skewed:  the differences between leader ratings and follower ratings were equally likely to be between -1 (follower one point higher than leader) and +2 (leader two points higher than follower).  

\begin{table}
    \centering
\caption{Overall and Per-Trial Subjective Fluency Ratings by Role and Interaction Type (\textbf{H}uman \textbf{H}uman \textbf{I}nteraction or \textbf{H}uman \textbf{R}obot \textbf{I}nteraction)}
    \begin{tabular}{c|c|c|c}
    Role, Interaction & Mean & Median & Mode \\
    \midrule
    Leader, HHI   &  4.294 & 5 & 5\\
    Follower, HHI   & 3.706 & 4 & 4\\
    Follower, HRI  & 3.313 & 3 & 4 \\
    Leader, HRI   &  2.627 & 3 & 2\\
    \end{tabular}
    \label{tab:fluencyratings}
\end{table}

\subsection{Haptic Collaboration Previews}
Of the 153 trials (17 pairs of participants each with three human-human, three human-leading-robot, and three robot-leading-human trials), 13 were excluded from our analysis due to missing/corrupted video recording (9) or because the leader did not follow the prescribed motion plan correctly (4).  From the remainder, we plot the linear acceleration by segment divided by various task factors.   

\begin{figure}
    \centering
    \begin{subfigure}[t]{\linewidth}
\includegraphics[width=3in]{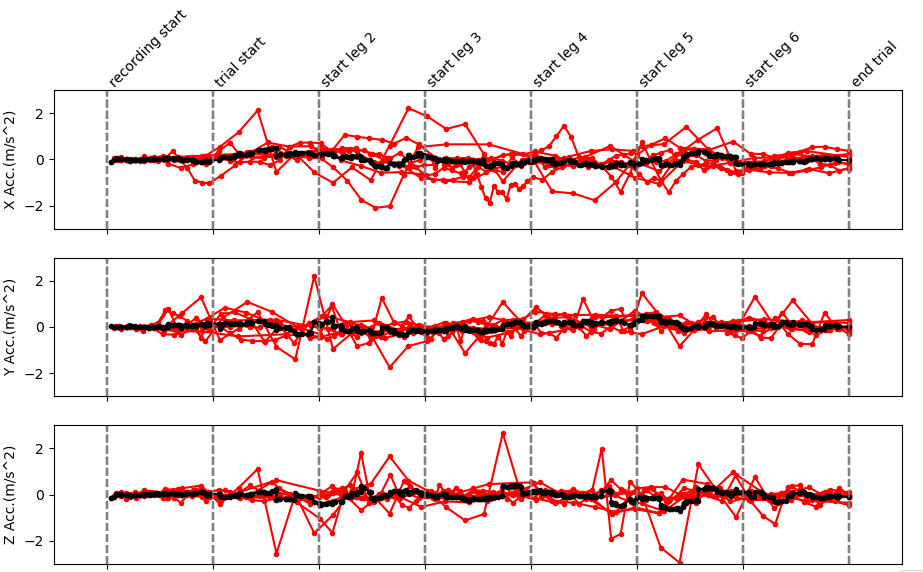}
    \caption{Human Human, M card, 9 groups and mean}
    \label{fig:IMUhhM}
    \end{subfigure}
    \begin{subfigure}[t]{\linewidth}
\includegraphics[width=3in]{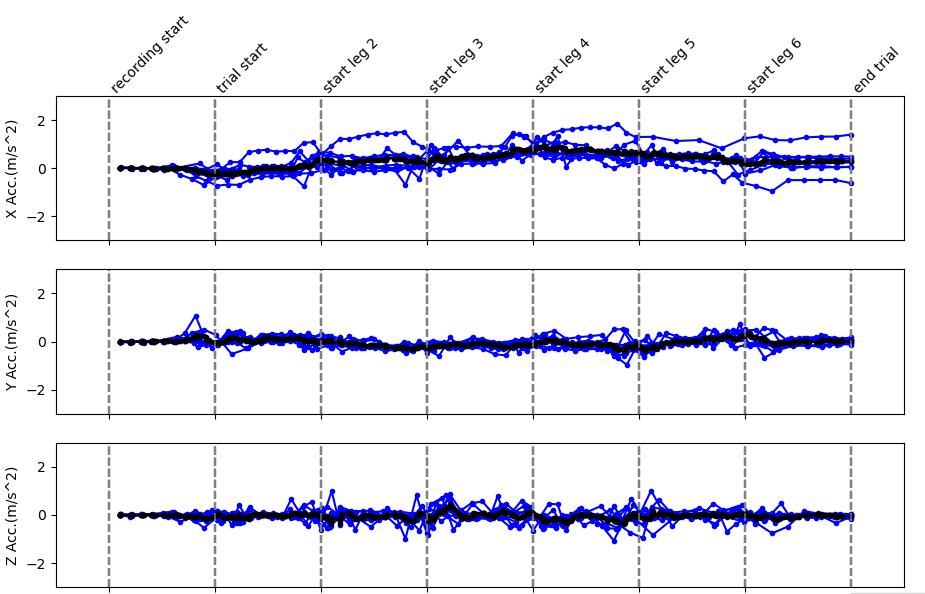}
    \caption{Robot Leading Human, M card, 8 groups and mean}
    \label{fig:IMUrleaderM}
    \end{subfigure}
\begin{subfigure}[t]{\linewidth}
\includegraphics[width=3in]{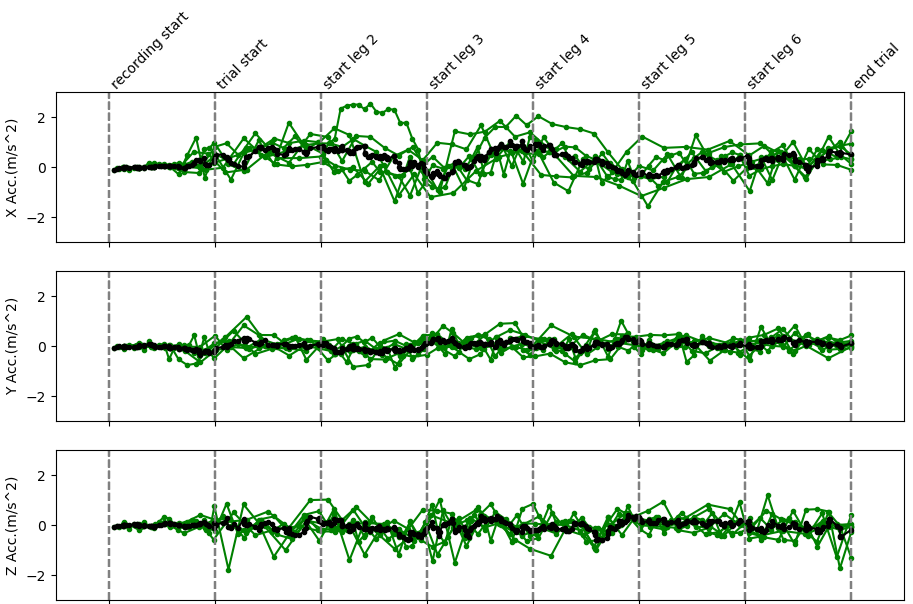}
    \caption{Human Leading Robot, M card, 7 groups and mean}
    \label{fig:IMUhleaderM}
    \end{subfigure}
    \label{IMUmultiplot}
    \caption{IMU data in three axes for all trials of a single card}
\end{figure}

We observe visual differences in the rates of change of acceleration of the three types of interactions, including acceleration prior to motion breakpoints (grey vertical lines separating segments).  Figure \ref{fig:humanhumanfluency5} has rounded ``hills'' and ``valleys'' of acceleration changes, while the robot-led, pre-programmed trajectory of \ref{fig:robotleaderfluency5} has sharp transitions and short bursts of acceleration that correspond with sudden initiation and cessation of robot movements.  While there are some acceleration profiles that seem to preview the motion of the following segment in the human-human collaborations, they are not clearly correlated or uniformly present before each segment boundary, even in the most fluent human-human collaborations.   

\subsection{Exploratory Analysis}
\subsubsection{Task Completion Time}
\begin{figure}
    \centering
    \includegraphics[width=\linewidth]{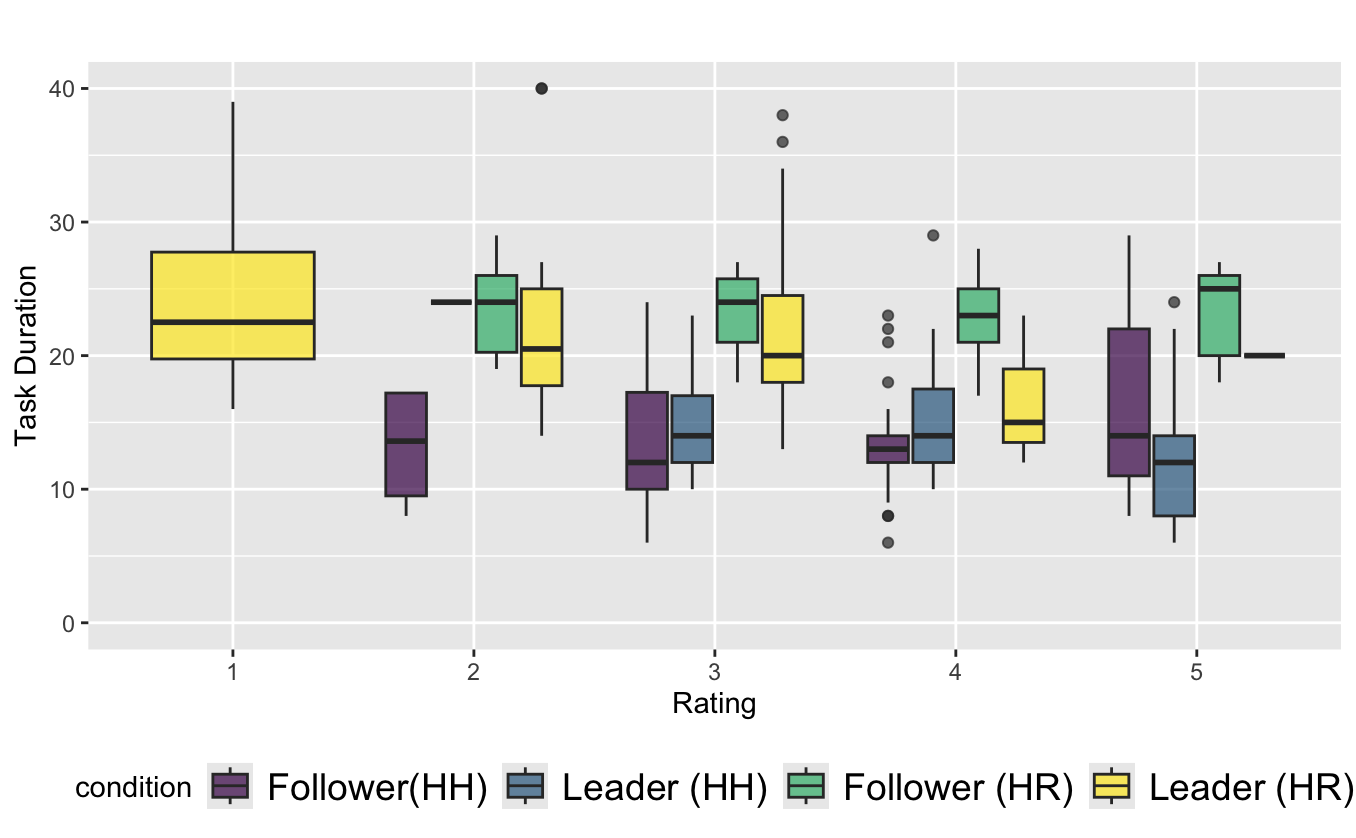}
    \caption{Task Time by Fluency Rating}
    \label{fig:ratingsvstime}
\end{figure}

Task completion time is frequently used as a proxy for collaboration fluency in human-robot interactions.  However, in human-human interactions in our task, we found that task completion time did not correlate with subjective evaluations of fluency. The distribution of task times for tasks rated as more or less fluent is shown in Figure \ref{fig:ratingsvstime}.  Pearson's product-moment correlation between task-specific rating and task time was found to be not significant (p=.39, correlation factor of -0.086) indicating that there is no correlation between task time and fluency for our human-human tasks.  

For our human-robot tasks with a human leader\footnote{We exclude robot-led tasks from this analysis as all the robot-led tasks were pre-planned and therefore have the same task completion time.}, the correlation of task time to fluency rating is again not straightforward.  Figure \ref{fig:ratingsvstime} shows the average time for human-led human-robot co-manipulations (yellow) and human-human co-manipulations (purple and blue). There is no significant correlation between time and fluency for human-human tasks (Spearman correlation -0.1, p=0.28) or human-robot tasks (Spearman correlation -0.04, p=0.66). While the lowest ratings (1) and the longest times (40 seconds) were both observed in the human leading robot condition, they did not occur in the same trial and the correlation is not statistically significant (Spearman correlation -0.24, p=0.08).

\subsubsection{Force and Jerk}
\begin{figure}
    \centering
    \begin{subfigure}[t]{\linewidth}
\includegraphics[height=2.5in]{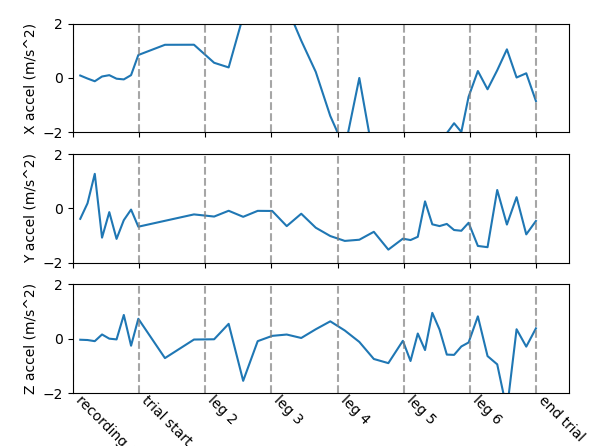}
    \caption{Human Leading Robot, Fluency 1}
    \label{fig:fluency1}
    \end{subfigure}
    \begin{subfigure}[t]{\linewidth}
\includegraphics[height=2.5in]{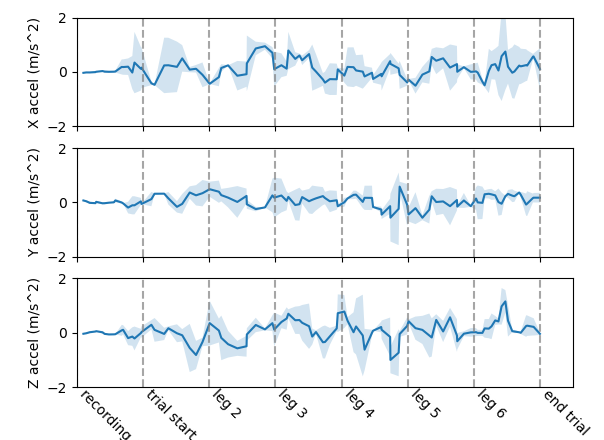}
    \caption{Human Leading Robot, Fluency 4}
    \label{fig:fluency4}
    \end{subfigure}
    \caption{IMU data from interactions grouped by subjective fluency ratings}
\end{figure}

During the experiment, we observed that some participants leading the robot appeared to have more difficulty, while others were able to lead the robot effectively.  The most visible feature of these difficult collaborations were oscillatory behaviors as the participant started a motion, did not observe the robot responding, and exerted more force resulting in an overshoot of the planned motion and the necessity for a corrective action.  (Figure \ref{fig:fluency1} shows an example of this behavior).  This oscillatory behavior is a known and undesirable feature of many human-robot shared control systems.  

%There are four potential fluency measures within this type of interaction. The first two are the magnitude of forces at the end effector of the robot (in any direction) and the \textbf{change} in forces experienced over a short duration (essentially the derivative of the force).  When the robot is in motion, these forces will be isomorphic to the acceleration data of the IMU and its derivative (jerk), which has the advantage of being directly comparable between the human-human and human-robot interactions. However, these metrics may not perfectly align in cases where the human leader is exerting force on the robot but it is resisting the desired motion---in this case, the robot end effector may be experiencing significant force while the IMU on the jointly-manipulated object would experience 0 acceleration.  
Because our subjective fluency ratings are only at the task level (we do not have instantaneous fluency ratings at intermediate points during the task), we compare the average IMU acceleration and average IMU acceleration change per timestep. To allow for analysis of the human-human interactions as well as the human-robot interactions, we focus on just the IMU data and not the robot end-effector data.

\begin{figure}
    \centering
    \includegraphics[width=0.9\linewidth]{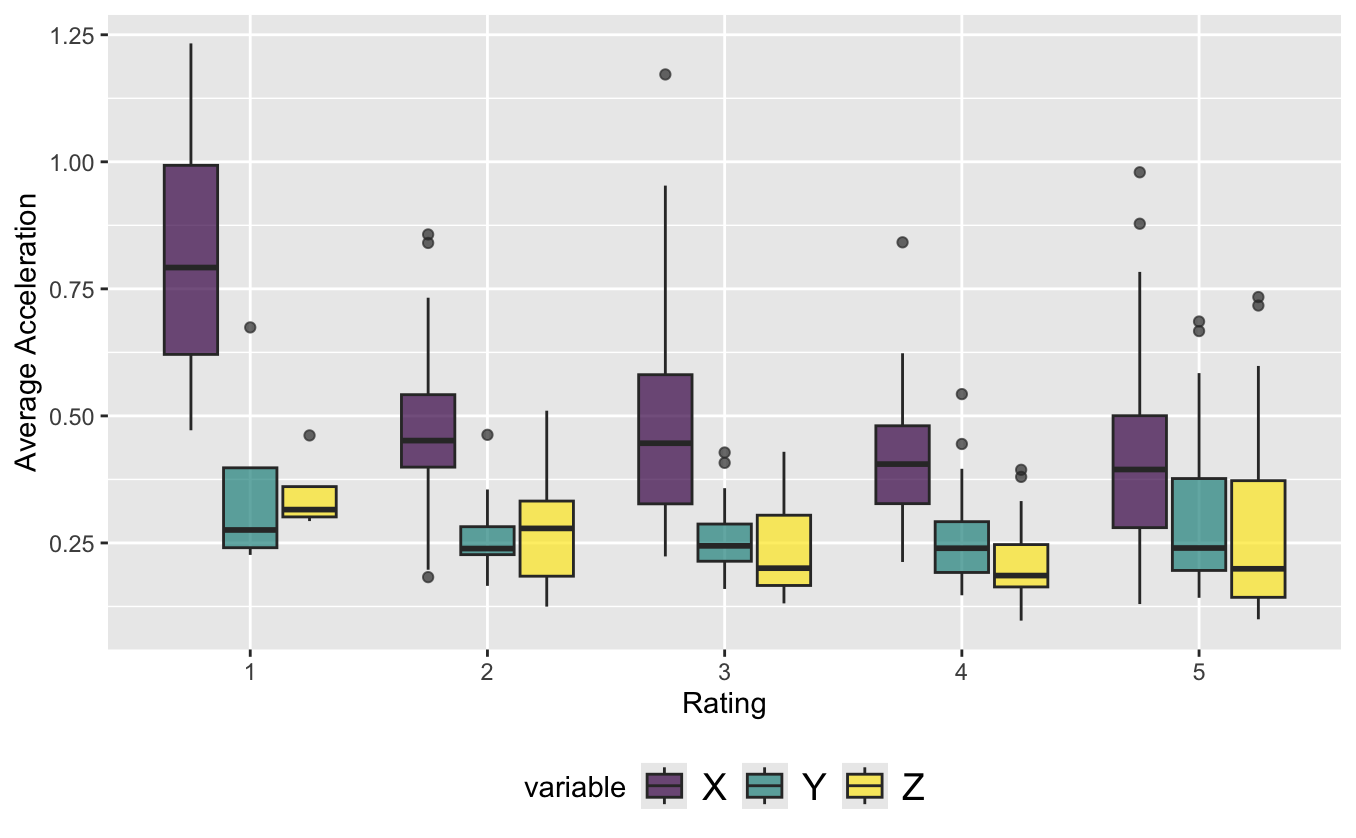}
    \caption{Mean Acceleration by Fluency Rating}
    \label{fig:meanaccelfluency}
\end{figure}

\begin{figure}
    \centering
    \includegraphics[width=\linewidth]{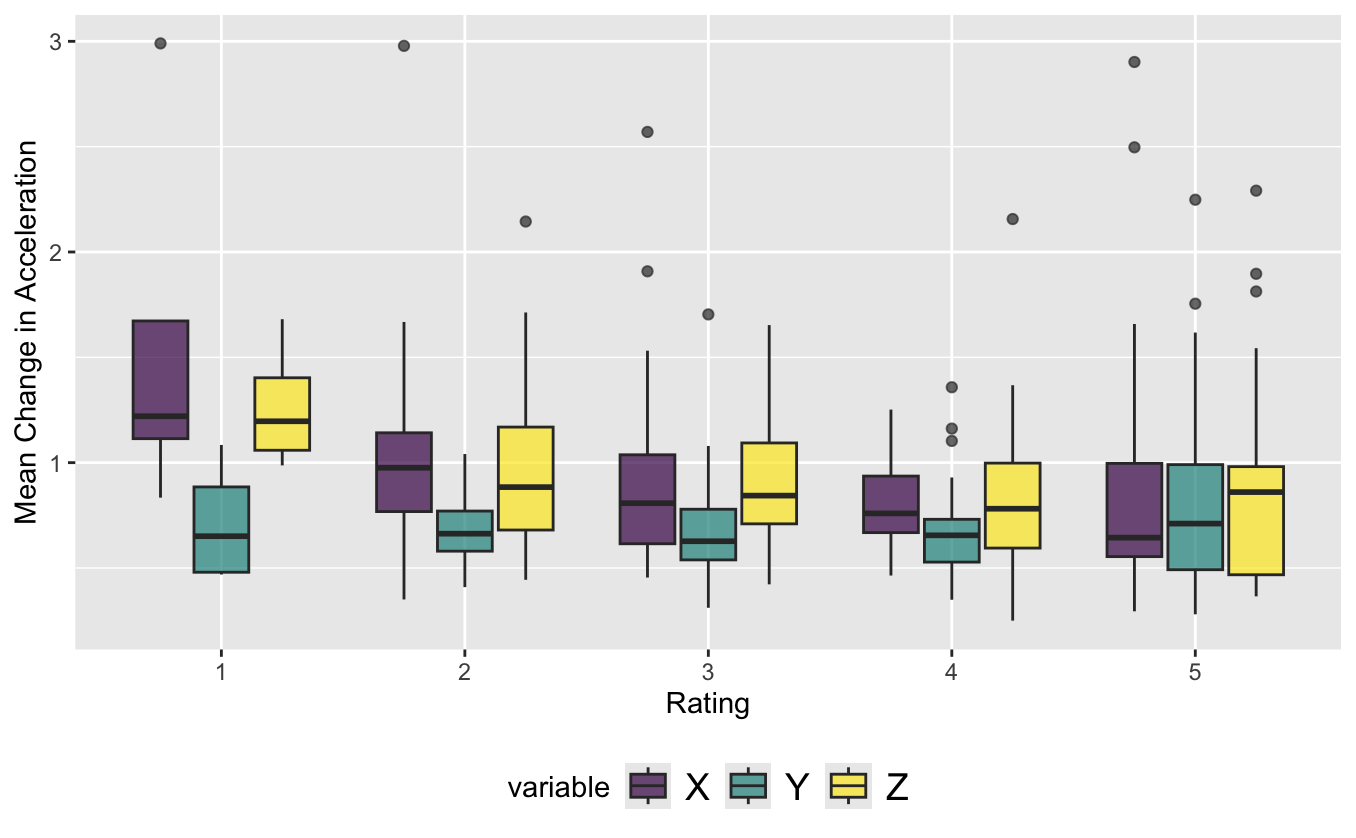}
    \caption{Mean Change in Acceleration by Fluency Rating}
    \label{fig:meandiffaccel}
\end{figure}

Fluency does have a correlation with average single-trial acceleration (Figure \ref{fig:meanaccelfluency}) and with average \emph{change} in acceleration between adjacent IMU datapoints (Figure \ref{fig:meandiffaccel}).  Average acceleration over the duration of a single trial is more than three times as high for the lowest-fluency trials as for the highest-fluency trials, with a Spearman correlation of -0.1 (p=0.039).  The maximum acceleration over the trial is also correlated with fluency, with a Spearman correlation of -0.1 (p=0.037).
The change in acceleration over the trial (point-by-point derivative) is correlated with fluency in all three axes with a Spearman correlation of -0.133 (p=0.006), but the \emph{maximum} change in acceleration is not significantly correlated with fluency (Spearman correlation of -0.049, p=0.309).

%\subsection{Fluency and Perception}

\section{Discussion}
\subsection{Haptic Communication Features}
 While the IMU acceleration data does show characteristic differences between the motion of human-human, human leading robot, and robot leading human motion trials, it is not clear from this dataset that this characteristic motion includes reliably identifiable motion previews distinguishable from noise, failing to prove our hypothesis \textbf{H2}.

\begin{figure}
% generated with python3 ~/Documents/GitHub/hapticcomm/processing/plotIMU-pandas.py . ../legtimes.csv in directory /Users/katallen/Dropbox/Research/HapticComm/Analysis/IMU_files/singletrials with options group8|group7,pentagon and all other options default
\centering
    \includegraphics[width=\linewidth]{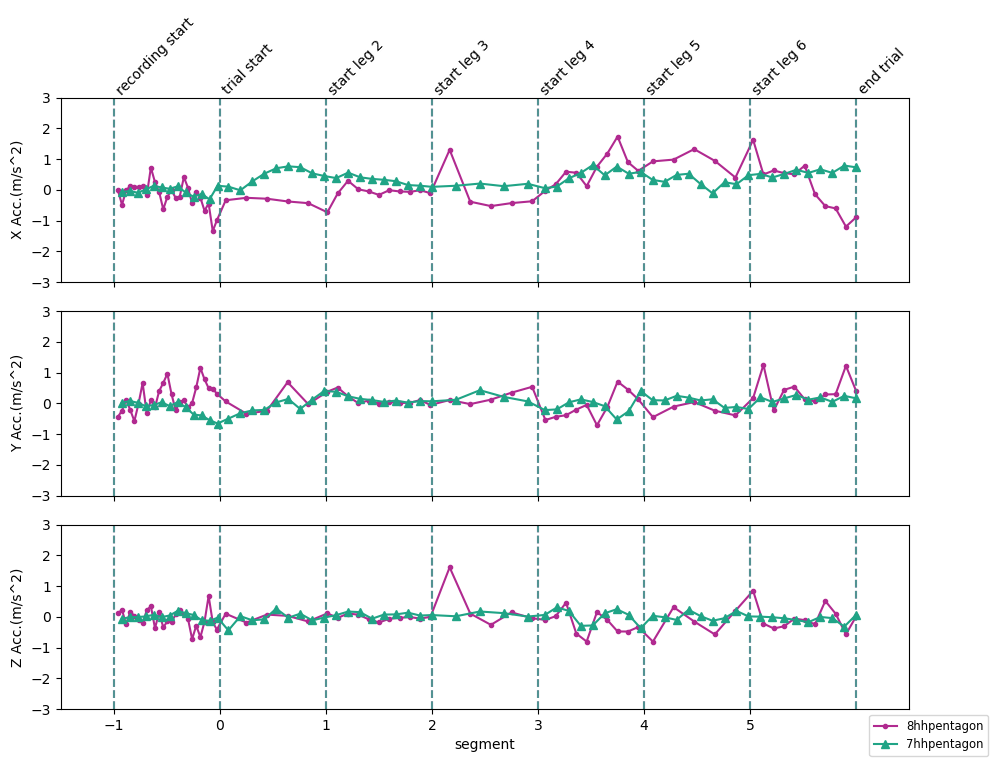}
    \caption{Two ``pentagon'' motion plans, both executed by human-human dyads.  Group 8 (rated as a fluency of 3 by both participants, in magenta with circle point markers) has a path with many apparently noisy features, while Group 7s path is smooth (rated a fluency of 5 by both participants, in teal with triangle point markers)}
    \label{fig:compare}
\end{figure}

However, the IMU data does show observable differences between human-human and human-robot acceleration profiles, and between profiles rated more or less fluent.  For example, in Figure \ref{fig:compare}, the very smooth profile of Group 7 was rated as much more fluent than the noisy profile of Group 8, suggesting that smooth changes in acceleration might be a feature of fluent co-manipulation.  These smooth motion profiles do appear more frequently in collaborations identified as fluent (partly supporting our hypothesis \textbf{H3}), as we observed in our exploratory analysis of technical metrics for fluency---visually smooth acceleration changes will result in numerically low mean changes in acceleration, one of our proposed technical metrics for fluency.

\subsection{Fluency Metrics}
In this experiment, we identified no correlation between fluency and time, which is commonly used as a proxy for subjective fluency. This suggests that task duration may not be an appropriate metric for fluency in this type of task.

We propose as alternatives average acceleration or average change in acceleration (Figures \ref{fig:meanaccel} and \ref{fig:meandiffaccel}), both of which correlate with our measures of subjective fluency.  

%\subsection{Fluency and Perception}

\subsection{Limitations and Future Work}
\label{futurework}
Our analysis is limited by several technical parameters of the experiment.  First, while the IMU data is essential to capture human-human interactions, the limitations of the low-cost IMU in both data rate and timestamping capability limit the data on which to base conclusions about signaling in various interactions, and make correlating precise clock-time events to the IMU data challenging.  A higher-fidelity sensor may be able to better distinguish between accidental and intentional motions to identify true haptic signaling, and timestamped data collection will allow for more precise correlation with video data.   Additional sensors on the wrists or arms of the participants may be necessary to identify and distinguish Leader-initiated motion from Follower-initiated motion of the co-manipulated object.

\section{Conclusion}
Haptic communication is used intuitively in physical collaboration tasks to communicate between human agents about safety, readiness, and to coordinate the group action.  For robotic participants to join physical collaborations as full participants, they require explicit training or programming to be able to interpret and identify these signals, and to generate them to coordinate with other agents.

Our work identifies a preliminary metric (IMU data) for identifying haptic communication in a physical collaboration, and demonstrates that haptic communication between  collaborators allows them to more effectively and efficiently complete their tasks, with higher satisfaction among the collaborators.

\addtolength{\textheight}{-12cm}   % This command serves to balance the column lengths

%%%%%%%%%%%%%%%%%%%%%%%%%%%%%%%%%%%%%%%%%%%%%%%%%%%%%%%%%%%%%%%%%%%%%%%%%%%%%%%%

%%%%%%%%%%%%%%%%%%%%%%%%%%%%%%%%%%%%%%%%%%%%%%%%%%%%%%%%%%%%%%%%%%%%%%%%%%%%%%%%

%%%%%%%%%%%%%%%%%%%%%%%%%%%%%%%%%%%%%%%%%%%%%%%%%%%%%%%%%%%%%%%%%%%%%%%%%%%%%%%%

%%%%%%%%%%%%%%%%%%%%%%%%%%%%%%%%%%%%%%%%%%%%%%%%%%%%%%%%%%%%%%%%%%%%%%%%%%%%%%%%

\bibliographystyle{IEEEtran}
\bibliography{haptics}

\begin{thebibliography}{10}
\providecommand{\url}[1]{#1}
\csname url@rmstyle\endcsname
\providecommand{\newblock}{\relax}
\providecommand{\bibinfo}[2]{#2}
\providecommand\BIBentrySTDinterwordspacing{\spaceskip=0pt\relax}
\providecommand\BIBentryALTinterwordstretchfactor{4}
\providecommand\BIBentryALTinterwordspacing{\spaceskip=\fontdimen2\font plus
\BIBentryALTinterwordstretchfactor\fontdimen3\font minus
  \fontdimen4\font\relax}
\providecommand\BIBforeignlanguage[2]{{%
\expandafter\ifx\csname l@#1\endcsname\relax
\typeout{** WARNING: IEEEtran.bst: No hyphenation pattern has been}%
\typeout{** loaded for the language `#1'. Using the pattern for}%
\typeout{** the default language instead.}%
\else
\language=\csname l@#1\endcsname
\fi
#2}}

\bibitem{reed_physical_2008}
K.~Reed and M.~Peshkin, ``Physical {Collaboration} of {Human}-{Human} and
  {Human}-{Robot} {Teams},'' \emph{Haptics, IEEE Transactions on}, vol.~1, pp.
  108--120, July 2008.

\bibitem{parker_design_2012}
C.~Parker and E.~Croft, ``Design \& {Personalization} of a {Cooperative}
  {Carrying} {Robot} {Controller},'' \emph{2012 IEEE International Conference
  on Robotics and Automation}, 2012.

\bibitem{dragan_effects_2015}
\BIBentryALTinterwordspacing
A.~D. Dragan, S.~Bauman, J.~Forlizzi, and S.~S. Srinivasa, ``Effects of {Robot}
  {Motion} on {Human}-{Robot} {Collaboration},'' in \emph{Proceedings of the
  {Tenth} {Annual} {ACM}/{IEEE} {International} {Conference} on {Human}-{Robot}
  {Interaction}}, ser. {HRI} '15.\hskip 1em plus 0.5em minus 0.4em\relax New
  York, NY, USA: Association for Computing Machinery, Mar. 2015, pp. 51--58.
  [Online]. Available: \url{http://doi.org/10.1145/2696454.2696473}
\BIBentrySTDinterwordspacing

\bibitem{al-saadi_novel_2021}
\BIBentryALTinterwordspacing
Z.~Al-Saadi, D.~Sirintuna, A.~Kucukyilmaz, and C.~Basdogan,
  ``\BIBforeignlanguage{en}{A {Novel} {Haptic} {Feature} {Set} for the
  {Classification} of {Interactive} {Motor} {Behaviors} in {Collaborative}
  {Object} {Transfer}},'' \emph{\BIBforeignlanguage{en}{IEEE Transactions on
  Haptics}}, vol.~14, no.~2, pp. 384--395, Apr. 2021. [Online]. Available:
  \url{https://ieeexplore.ieee.org/document/9241412/}
\BIBentrySTDinterwordspacing

\bibitem{rysbek_recognizing_2023}
\BIBentryALTinterwordspacing
Z.~Rysbek, K.~H. Oh, and M.~Zefran, ``\BIBforeignlanguage{en}{Recognizing
  {Intent} in {Collaborative} {Manipulation}},'' in
  \emph{\BIBforeignlanguage{en}{{INTERNATIONAL} {CONFERENCE} {ON} {MULTIMODAL}
  {INTERACTION}}}, Oct. 2023, pp. 498--506, arXiv:2308.09177 [cs]. [Online].
  Available: \url{http://arxiv.org/abs/2308.09177}
\BIBentrySTDinterwordspacing

\bibitem{yang_implicit_2025}
E.~Yang and C.~Mavrogiannis, ``Implicit {Communication} in {Human}-{Robot}
  {Collaborative} {Transport},'' in \emph{Proceedings of the 2025 {ACM}/{IEEE}
  {International} {Conference} on {Human}-{Robot} {Interaction}}, ser. {HRI}
  '25.\hskip 1em plus 0.5em minus 0.4em\relax Melbourne, Australia: IEEE Press,
  Mar. 2025, pp. 23--33.

\bibitem{carpendale_development_2010}
\BIBentryALTinterwordspacing
J.~I.~M. Carpendale and A.~B. Carpendale, ``\BIBforeignlanguage{english}{The
  {Development} of {Pointing}: {From} {Personal} {Directedness} to
  {Interpersonal} {Direction}},'' \emph{\BIBforeignlanguage{english}{Human
  Development}}, vol.~53, no.~3, pp. 110--126, 2010, publisher: Karger
  Publishers. [Online]. Available:
  \url{https://www.karger.com/Article/FullText/315168}
\BIBentrySTDinterwordspacing

\bibitem{rosenbaum_cognition_2012}
D.~A. Rosenbaum, K.~M. Chapman, M.~Weigelt, D.~J. Weiss, and R.~van~der Wel,
  ``Cognition, action, and object manipulation,'' \emph{Psychological
  Bulletin}, vol. 138, no.~5, pp. 924--946, 2012, place: US Publisher: American
  Psychological Association.

\bibitem{pezzulo_human_2013}
\BIBentryALTinterwordspacing
G.~Pezzulo, F.~Donnarumma, and H.~Dindo, ``\BIBforeignlanguage{en}{Human
  {Sensorimotor} {Communication}: {A} {Theory} of {Signaling} in {Online}
  {Social} {Interactions}},'' \emph{\BIBforeignlanguage{en}{PLOS ONE}}, vol.~8,
  no.~11, p. e79876, Nov. 2013, publisher: Public Library of Science. [Online].
  Available:
  \url{https://journals.plos.org/plosone/article?id=10.1371/journal.pone.0079876}
\BIBentrySTDinterwordspacing

\bibitem{admoni_deliberate_2014}
\BIBentryALTinterwordspacing
H.~Admoni, A.~Dragan, S.~S. Srinivasa, and B.~Scassellati, ``Deliberate delays
  during robot-to-human handovers improve compliance with gaze communication,''
  in \emph{Proceedings of the 2014 {ACM}/{IEEE} international conference on
  {Human}-robot interaction}, ser. {HRI} '14.\hskip 1em plus 0.5em minus
  0.4em\relax New York, NY, USA: Association for Computing Machinery, Mar.
  2014, pp. 49--56. [Online]. Available:
  \url{https://doi.org/10.1145/2559636.2559682}
\BIBentrySTDinterwordspacing

\bibitem{chan_grip_2012}
\BIBentryALTinterwordspacing
W.~P. Chan, C.~A. Parker, H.~M. Van~der Loos, and E.~A. Croft, ``Grip forces
  and load forces in handovers: implications for designing human-robot handover
  controllers,'' in \emph{Proceedings of the seventh annual {ACM}/{IEEE}
  international conference on {Human}-{Robot} {Interaction}}, ser. {HRI}
  '12.\hskip 1em plus 0.5em minus 0.4em\relax New York, NY, USA: Association
  for Computing Machinery, Mar. 2012, pp. 9--16. [Online]. Available:
  \url{https://doi.org/10.1145/2157689.2157692}
\BIBentrySTDinterwordspacing

\bibitem{parker_experimental_2011}
C.~A.~C. Parker and E.~A. Croft, ``Experimental investigation of human-robot
  cooperative carrying,'' in \emph{2011 {IEEE}/{RSJ} {International}
  {Conference} on {Intelligent} {Robots} and {Systems}}, Sept. 2011, pp.
  3361--3366, iSSN: 2153-0866.

\bibitem{aydin_computational_2020}
\BIBentryALTinterwordspacing
Y.~Aydin, O.~Tokatli, V.~Patoglu, and C.~Basdogan, ``A {Computational}
  {Multicriteria} {Optimization} {Approach} to {Controller} {Design} for
  {Physical} {Human}-{Robot} {Interaction},'' \emph{IEEE Transactions on
  Robotics}, vol.~36, no.~6, pp. 1791--1804, Dec. 2020. [Online]. Available:
  \url{https://ieeexplore.ieee.org/document/9162045}
\BIBentrySTDinterwordspacing

\end{thebibliography}

\end{document}